\documentclass[10pt,twocolumn,letterpaper]{article}
\usepackage[pagenumbers]{cvpr}
\usepackage{graphicx}
\usepackage{amsmath,amssymb,amsfonts}
\usepackage{textcomp}
\usepackage{makecell}
\usepackage{colortbl}
\usepackage{xcolor}
\usepackage{booktabs}
\usepackage{multirow}
\usepackage{tabularx}
\usepackage{array}
\usepackage[breaklinks,colorlinks]{hyperref}
\hypersetup{pdftitle={SalivaryHSI-Seg: A New Challenging Benchmark for Ex Vivo Salivary Gland Hyperspectral Tissue Segmentation}, pdfauthor={Xiaoqi Guo, Xianran Wang, Xinquan Yang, Qinyi Lu, Fangqian Zhang, Jiangyi Chen, Ming Li, Wen Ma, Linlin Shen}}

\newcommand{\cmark}{\ensuremath{\checkmark}}
\newcommand{\pmark}{\ensuremath{\triangle}}

\newcolumntype{L}[1]{>{\raggedright\arraybackslash}p{#1}}
\newcolumntype{C}[1]{>{\centering\arraybackslash}p{#1}}
\newcolumntype{Y}{>{\raggedright\arraybackslash}X}

\begin{document}
\title{SalivaryHSI-Seg: A New Challenging Benchmark for Ex Vivo Salivary Gland Hyperspectral Tissue Segmentation}
\author{Xiaoqi Guo$^{1,*}$, Xianran Wang$^{2,*}$, Xinquan Yang$^{1}$, Qinyi Lu$^{2}$, Fangqian Zhang$^{2}$, \\
Jiangyi Chen$^{2}$, Ming Li$^{2}$, Wen Ma$^{2,\dagger}$, Linlin Shen$^{1,\dagger}$ \\
$^1$College of Computer Science and Software Engineering and School of Artificial Intelligence, \\
Shenzhen University, Shenzhen, China \\
$^2$Department of Oral and Maxillofacial Surgery, Digital Center of Stomatology, \\
The Affiliated Stomatology Hospital, Kunming Medical University, China \\
{\small 2022110135@email.szu.edu.cn; llshen@szu.edu.cn; mawen@kmmu.edu.cn} \\
{\small $^*$Equal contribution. $^\dagger$Corresponding authors.}}
\maketitle

\begin{abstract}
Salivary gland tumors are common lesions in oral and maxillofacial practice, and accurate identification of tumor regions and surrounding salivary gland tissues is important for tissue assessment, pathological sampling, and clinical decision support. Hyperspectral imaging (HSI) records continuous spectral responses together with spatial imaging information, providing a promising way to capture fine-grained tissue-discriminative information. Nevertheless, the development of salivary gland HSI tissue segmentation remains limited due to the lack of standardized benchmarks with pixel-level annotations and unified evaluation protocols. To address this issue, we introduce SalivaryHSI-Seg, a hyperspectral tissue segmentation benchmark for ex vivo salivary gland specimens, containing 119 HSI samples from 31 patients with multi-class pixel-level annotations. We establish a patient-level five-fold cross-validation protocol and evaluate different input representations, model architectures, and challenging subsets with small tumor and complex boundary. Based on this benchmark, we further
propose HyBSD, a hyperspectral boundary-guided spectral-difference network for full-HSI salivary gland tissue segmentation. HyBSD exploits boundary-guided spectral-difference cues to improve tissue delineation in ambiguous transition regions. Extensive experiments show that full-HSI input provides more discriminative information than sampled spectrums, and that HyBSD achieves 74.97\% mDice and 87.76\% Tumor Dice, with further advantages on the challenging subsets. Dataset and code will be released.
\end{abstract}


\section{Introduction}
\label{sec:introduction}
Salivary gland tumors are common lesions in oral and maxillofacial practice, among which pleomorphic adenoma (PA) and Warthin tumor (WT) are two representative benign tumor types. Accurate identification of tumor regions and surrounding salivary gland tissues is important for tissue assessment, pathological sampling, and clinical decision support. However, ex vivo salivary gland specimens often contain mixed tissue components, including glandular parenchyma, connective tissue septa, and tumor regions, making fine-grained tissue distinction challenging. In routine gross observation or RGB imaging, tissue assessment mainly relies on visible color and morphological information, but these appearance cues may not sufficiently reflect subtle differences among different tissue components. Especially in transition regions between tumor and non-tumor tissues, different tissues may exhibit similar visual appearances and gradually changing boundaries, making RGB-based observation insufficient for reliably distinguishing tissue boundaries.

In recent years, hyperspectral imaging (HSI) has been increasingly explored for medical tissue analysis~\cite{lu2014medical,tran2026medical}, including tissue recognition~\cite{studier2022spectral}, lesion detection~\cite{leon2023hyperspectral}, and intraoperative tissue segmentation~\cite{roddan2025samsa}. HSI records a continuous reflectance spectrum for each pixel together with two-dimensional spatial imaging, forming a three-dimensional data cube with both spatial and spectral dimensions. Compared with conventional RGB imaging, HSI captures tissue morphology while preserving subtle spectral differences related to biochemical and structural tissue properties~\cite{studier2022spectral}, making it a promising imaging technique for fine-grained salivary gland tumor tissue segmentation.

Despite the potential of HSI for salivary gland tumor tissue recognition, publicly available salivary gland tumor HSI datasets with pixel-level tissue annotations remain very limited. Such annotations require substantial manual effort and pathological knowledge. %Existing public salivary gland tumor datasets mainly target molecular profiling, radiological analysis, or digital pathology analysis, and cannot directly support ex vivo salivary gland HSI tissue segmentation. 
As summarized in Table~\ref{tab:dataset_comparison}, existing human
medical HSI datasets cover a variety of anatomical targets, but most
lack complete and dense pixel-level semantic annotations. Meanwhile,
no publicly available HSI dataset currently provides dense pixel-level
semantic annotations specifically for ex vivo salivary gland tissue
segmentation. Existing public salivary gland tumor datasets, on the
other hand, mainly focus on molecular profiling, radiological analysis,
or digital pathology, and therefore cannot be directly applied to this task.
Therefore, the community needs a standardized benchmark with pixel-level annotations and a unified evaluation protocol.

To this end, we construct SalivaryHSI-Seg, a medical hyperspectral tissue segmentation benchmark for ex vivo salivary gland specimens. The dataset contains 119 HSI samples from 31 patients, including pleomorphic adenoma (PA), Warthin tumor (WT), and normal samples. Each sample contains 256 continuous spectral bands covering approximately 386--1011 nm, with the effective range mainly spanning 400--1000 nm. SalivaryHSI-Seg provides pixel-level tissue annotations for background, glandular parenchyma, connective tissue septa, PA tumor, and WT tumor. The annotations were independently delineated on HSI-derived pseudo-RGB images by two clinicians with more than 3 and 5 years of clinical experience, respectively, with disagreements adjudicated by an experienced oral histopathologist. To support systematic evaluation, we adopt a patient-level five-fold cross-validation protocol, compare different input representations and segmentation architectures, and define two challenge subsets, whose samples contain small tumor and complex boundary, to evaluate model behavior beyond overall average performance.

% Dataset comparison table
\begin{table*}[!t]
\centering
\caption{Comparison of representative medical HSI datasets
for human tissue analysis.}
\label{tab:dataset_comparison}

\footnotesize
\setlength{\tabcolsep}{2.3pt}
\renewcommand{\arraystretch}{1.18}

\resizebox{\textwidth}{!}{%
\begin{tabular}{
@{}
L{3.55cm}
C{2.05cm}
C{1.25cm}
C{0.95cm}
C{2.75cm}
C{1.45cm}
C{0.75cm}
C{1.35cm}
C{1.45cm}
C{0.70cm}
@{}
}
\toprule

\multicolumn{1}{c}{\multirow{2}{*}{\textbf{Dataset}}} &
\multicolumn{2}{c}{\textbf{Acquisition}} &
\multicolumn{2}{c}{\textbf{Spectral Information}} &
\multicolumn{2}{c}{\textbf{Dataset Size}} &
\multicolumn{3}{c}{\textbf{Annotations}} \\

\cmidrule(lr){2-3}
\cmidrule(lr){4-5}
\cmidrule(lr){6-7}
\cmidrule(lr){8-10}

&
\multicolumn{1}{c}{\textbf{Target}} &
\multicolumn{1}{c}{\textbf{Setting}} &
\multicolumn{1}{c}{\textbf{Bands}} &
\multicolumn{1}{c}{\textbf{Range (nm)}} &
\multicolumn{1}{c}{\textbf{Specimens}} &
\multicolumn{1}{c}{\textbf{HSI}} &
\multicolumn{1}{c}{\textbf{Pixel Mask}} &
\multicolumn{1}{c}{\textbf{Dense Labels}} &
\multicolumn{1}{c}{\textbf{Classes}} \\

\midrule

ODSI-DB~\cite{hyttinen2020oral} &
Oral/dental &
In vivo &
51; 204 &
\mbox{450--950; 400--1000} &
30 &
316 &
\pmark &
\textemdash &
35 \\

HELICoiD~\cite{fabelo2019vivo} &
Brain &
In vivo &
826 &
\mbox{400--1000} &
22 &
36 &
\pmark &
\textemdash &
4 \\

SLIMBRAIN~\cite{martin2025slimbrain} &
Brain &
In vivo &
25; 369 &
\mbox{660--950; 400--1000} &
193 &
193 &
\pmark &
\textemdash &
19 \\

Leipzig HSI~\cite{pathak2023spectral} &
Thyroid/GI &
In vivo &
100 &
\mbox{500--1000} &
100 &
NR &
\pmark &
\textemdash &
15 \\

Placenta HSI~\cite{puustinen2023hyperspectral} &
Placenta &
Ex vivo &
38--78 &
\mbox{515--700; 515--900} &
4 &
101 &
\cmark &
\pmark &
4 \\

GlioHyper~\cite{giannoni2024transportable} &
Brain glioma &
Ex vivo &
79 &
\mbox{510--900} &
11 &
12 &
\textemdash &
\textemdash &
3 \\

\midrule

\mbox{\textbf{SalivaryHSI-Seg (Ours)}} &
\mbox{\textbf{Salivary gland}} &
Ex vivo &
\textbf{256} &
\mbox{\textbf{386--1011}} &
\textbf{31} &
\textbf{119} &
\cmark &
\cmark &
\textbf{5} \\

\bottomrule
\end{tabular}%
}

\vspace{3pt}

\parbox{\textwidth}{%
\scriptsize
\textit{Note:} ``Classes'' denotes the number of annotated label categories
provided by each dataset. \checkmark indicates available annotations,
$\triangle$ denotes sparse, partial, or representative-region
annotations, and --- indicates unavailable annotations.
``Specimens'' and ``HSI'' report the numbers of human subjects or
tissue specimens and HSI acquisitions, respectively.
Semicolon-separated band numbers and spectral ranges correspond to
different HSI systems. NR denotes not reported.
}

\end{table*}

Beyond benchmark construction, we further observe that tissue transition regions in ex vivo salivary gland specimens often exhibit boundary ambiguity, making accurate tissue and tumor delineation more challenging. Meanwhile, local spectral variations around tissue interfaces provide useful cues for distinguishing adjacent tissue categories. Motivated by this observation, we design a hyperspectral boundary-guided spectral-difference network (HyBSD), which aims to learn spatial-spectral representations from the complete HSI cube.
Specifically, we introduce Boundary-Spectral Coupled Enhancement (BSCE), Semantic Boundary Supervision (SBS), and a Boundary-aware Refinement Branch (BAR). BSCE injects boundary-guided spectral-difference information into high-level 3D features, SBS provides category-related foreground boundary constraints, and BAR refines the base segmentation through residual logits correction.

The main contributions of this work are summarized as follows:
\begin{itemize}
    \item We construct and release SalivaryHSI-Seg, the first HSI dataset for ex vivo salivary gland tumor tissue segmentation, containing 119 HSI samples from 31 patients with expert-annotated multi-class pixel-level tissue labels.

    \item We propose HyBSD, a hyperspectral boundary-guided spectral-difference network for full-HSI salivary gland tissue segmentation. The network introduces Boundary-Spectral Coupled Enhancement (BSCE) to enhance boundary-guided spectral-difference cues, Semantic Boundary Supervision (SBS) to constrain category-related foreground boundaries, and a Boundary-aware Refinement Branch (BAR) to refine the base segmentation logits through residual correction.

    \item We establish a patient-level five-fold evaluation protocol and systematically evaluate medical HSI tissue segmentation from the perspectives of input spectrums, model architectures, and challenging subsets. Extensive experiments show that full-HSI input provides more discriminative information than reduced spectral representations, and that HyBSD achieves the best performance on overall metrics and difficult tumor-related subsets.
\end{itemize}

\section{Related Work}

%% 2.1 医学高光谱图像benchmark
\subsection{Hyperspectral Medical Image Datasets}

In recent years, several hyperspectral medical imaging datasets have been released to support algorithm development and evaluation across different clinical and biomedical scenarios~\cite{ali2025surgical}. ODSI-DB provides in-vivo hyperspectral reflectance images of the human oral and dental regions with heterogeneous expert annotations, and has been used for spectral tissue classification and partial-region segmentation studies~\cite{hyttinen2020oral,garcia2023hyperspectral}. HELICoiD-related brain HSI datasets~\cite{fabelo2019vivo,leon2023hyperspectral} and SLIMBRAIN~\cite{martin2025slimbrain} provide intraoperative human brain data with partial or sparse pixel-level annotations to support brain tissue identification and tumour analysis. The Leipzig human HSI cohort contains thyroid, parathyroid, and gastrointestinal tissue data acquired during open surgery and is primarily intended for discrimination among different tissue types~\cite{pathak2023spectral}. Fresh ex-vivo human tissue datasets have also been released, including the Hyperspectral Placenta Dataset for the analysis of placental vasculature and anatomical structures~\cite{puustinen2023hyperspectral}, and GlioHyper for biochemical mapping and preliminary grading of fresh glioma biopsy specimens~\cite{giannoni2024transportable}.

As shown in Table~\ref{tab:dataset_comparison}, existing medical HSI datasets differ substantially in anatomical target, acquisition setting, annotation granularity, and supported tasks. Most datasets provide sparse pixel annotations, selected regions of interest, masks of specific anatomical structures, or sample-level labels, rather than dense semantic annotations covering the complete tissue analysis region.

In contrast, SalivaryHSI-Seg provides dense and mutually exclusive pixel-level semantic annotations for fresh ex-vivo salivary-gland specimens. Its label taxonomy encompasses both normal tissue components and two clinically relevant tumour types, thereby supporting fine-grained multiclass semantic segmentation and model evaluation based on patient-level data splits.

\subsection{HSI Tissue Segmentation}

Medical HSI has also been investigated for dense tissue analysis in several segmentation-related tasks. In surgical scenes, deep models have been evaluated for hyperspectral organ segmentation, showing that spatial-spectral information can support semantic tissue recognition~\cite{seidlitz2022robust,roddan2024calibration}. Oral and dental spectral imaging studies have compared HSI, RGB, spectral-pixel, and spatial-spectral inputs for multi-class tissue segmentation~\cite{garcia2023hyperspectral,thiem2021hyperspectral}. HSI has also been applied to lymph node segmentation in oral squamous cell carcinoma~\cite{10381748}. In-vivo brain HSI studies have explored the discrimination of tumor tissue, normal tissue, blood vessels, and background for intraoperative brain tumor delineation~\cite{fabelo2019vivo}. In salivary gland-related HSI research, Halicek et al. investigated HSI-based tumor detection for thyroid and salivary gland specimens using deep learning~\cite{halicek2020tumor}. Other studies have investigated HSI-based segmentation or pixel-wise classification for tongue tumors~\cite{trajanovski2020tongue}, liver surgical specimens~\cite{zhang2021explainable}, pigmented skin lesions~\cite{aloupogianni2022hyperspectral}, and metastatic colon cancer frozen sections~\cite{kopriva2025hyperspectral}.

Recent studies have further explored Transformer-based, semi-supervised, and boundary-aware learning for medical HSI segmentation~\cite{yun2023spectr,qin2025semi,cao2025boundary}. More broadly, medical image segmentation studies have considered boundary-aware learning~\cite{KERVADEC2021101851,10946212} and probabilistic representations for uncertain regions~\cite{10723767}.

These HSI studies demonstrate the potential of HSI for medical tissue segmentation and tumor analysis, but existing tasks are often formulated as organ-level segmentation, binary lesion analysis, brain tumor delineation, tumor detection, or pathology-oriented pixel-wise classification. Different from these settings, SalivaryHSI-Seg focuses on fine-grained multi-class tissue segmentation within ex vivo salivary gland specimens, requiring the model to distinguish glandular parenchyma, connective tissue septa, PA tumor, and WT tumor at the pixel level.

%% 2.3  医学图像分割
%\subsection{Medical Image Segmentation}

%Medical image segmentation has been widely studied across different imaging modalities. U-Net-style encoder-decoder networks~\cite{ronneberger2015u}, 3D CNNs such as 3D U-Net~\cite{cciccek20163d}, and Transformer-based models such as UNETR~\cite{hatamizadeh2022unetr} and SwinUNETR~\cite{tang2022self} have become representative architectures for organ and lesion segmentation. These model families provide important baselines for evaluating medical HSI segmentation methods.

%However, most of them are designed for conventional 2D or 3D medical images, whose input assumptions differ from full-HSI data with high-dimensional continuous spectral responses~\cite{bench2022unsupervised}. Low-dimensional HSI representations may lose useful spectral cues, whereas full-HSI input requires effective spatial-spectral modeling under limited medical data. Therefore, we evaluate representative 2D CNN, 3D CNN, and Transformer-based segmentation models and further develop a boundary-aware full-HSI baseline for fine-grained salivary gland tissue segmentation.

%%------------数据集图-------------%%
\begin{figure}[!t]
\centering
\includegraphics[width=\columnwidth]{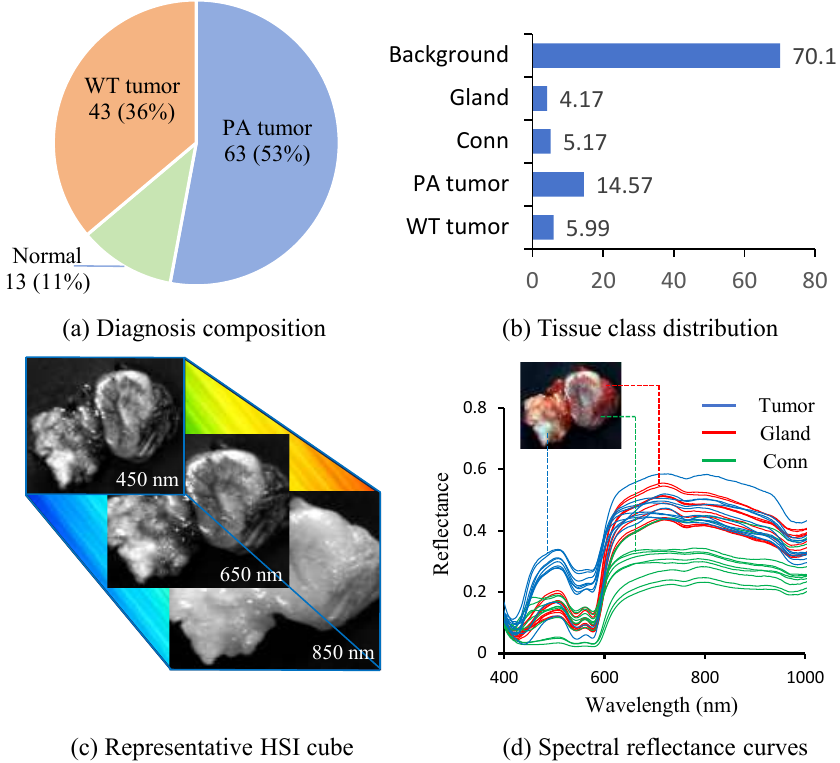}
\caption{Overview of SalivaryHSI-Seg. Tumor in the spectral curves denotes the diagnosis-specific tumor category in the selected sample.}
\label{fig:dataset}
\end{figure}

% SalivaryHSI-Seg 基准数据集
\section{The SalivaryHSI-Seg Benchmark}

% 3.1 数据采集与标注
\subsection{Data Acquisition and Annotation}

The data used in this study were collected from patients with salivary gland tumors who underwent surgical treatment, including pleomorphic adenoma (PA) and Warthin tumor (WT) confirmed by postoperative pathological examination. All patients provided informed consent, and this study was approved by the institutional ethics committee. Patient information was anonymized during data processing. Cases were included if they had a clear postoperative pathological diagnosis and complete clinical and pathological records. Cases with controversial diagnoses, incomplete records, or preoperative treatments that could potentially affect tissue spectral responses were excluded.

After surgical resection, ex vivo salivary gland specimens were immediately placed in sealed specimen bags, stored at room temperature, and subjected to hyperspectral imaging and pathological submission within 30 minutes. HSI data were acquired using a GaiaField-V10 hyperspectral imaging system in a darkroom environment. For each patient, one or more hyperspectral fields were acquired from the resected specimen according to specimen size and tissue availability; therefore, an HSI sample in SalivaryHSI-Seg denotes one acquired hyperspectral image rather than an individual patient. Two halogen lamps were used as illumination sources. The distance between the specimen and the camera was fixed at approximately 35 cm, and the field of view was 20$^\circ$. For each acquisition, white reference and dark reference images were collected for reflectance calibration:
\begin{equation}
R = \frac{I_{\mathrm{raw}} - I_{\mathrm{dark}}}
{I_{\mathrm{white}} - I_{\mathrm{dark}}}.
\label{eq:reflectance_calibration}
\end{equation}\textbf{}
The resulting HSI cube contains 256 continuous spectral bands, covering approximately 386--1011 nm, with the effective spectral range mainly spanning 400--1000 nm from the visible to near-infrared spectrum.

Pixel-level annotations were independently delineated on pseudo-RGB images synthesized from the HSI data by two clinicians, one with more than 3 years and the other with more than 5 years of clinical experience in the diagnosis and management of salivary gland lesions. Regions consistently identified by both clinicians were directly retained. Disagreements in tissue category or boundary delineation were adjudicated by an oral histopathologist with extensive diagnostic experience, which lead to the final annotation. During annotation, the annotators referred to gross-specimen pseudo-RGB images synthesized from the HSI data and determined tissue categories in conjunction with pathological examination results. The dataset contains five pixel-level tissue categories: background, glandular parenchyma, connective tissue septa, PA tumor, and WT tumor. Regions affected by specular reflection or other factors that hinder reliable interpretation were identified and excluded from training loss and metric calculation.

% 3.2 数据集统计与挑战属性
%\subsection{Dataset Statistics and Challenge Attributes}
\subsection{Dataset Statistics and Characteristics}

SalivaryHSI-Seg contains 119 ex vivo salivary gland HSI samples acquired from 31 patients, including 63 PA, 43 WT, and 13 Normal samples, where each sample corresponds to one hyperspectral acquisition field. Each sample was imaged with 256 continuous spectral bands, with image heights ranging from 196 to 1285 pixels and widths ranging from 191 to 1350 pixels. The dataset provides pixel-level annotations for background, glandular parenchyma, connective tissue septa, PA tumor, and WT tumor. Fig.~\ref{fig:dataset} provides an overview of SalivaryHSI-Seg.

%Within valid annotated regions, SalivaryHSI-Seg exhibits clear class imbalance. 
SalivaryHSI-Seg has an imbalanced class distribution. Background pixels account for 70.10\%, while glandular parenchyma and connective tissue septa account for 4.17\% and 5.17\%, respectively. PA tumor and WT tumor account for 14.57\% and 5.99\%, respectively. This distribution requires models to distinguish both tumor-related and non-tumor tissue categories.

\begin{figure}[t]
    \centering
    \includegraphics[width=\columnwidth]{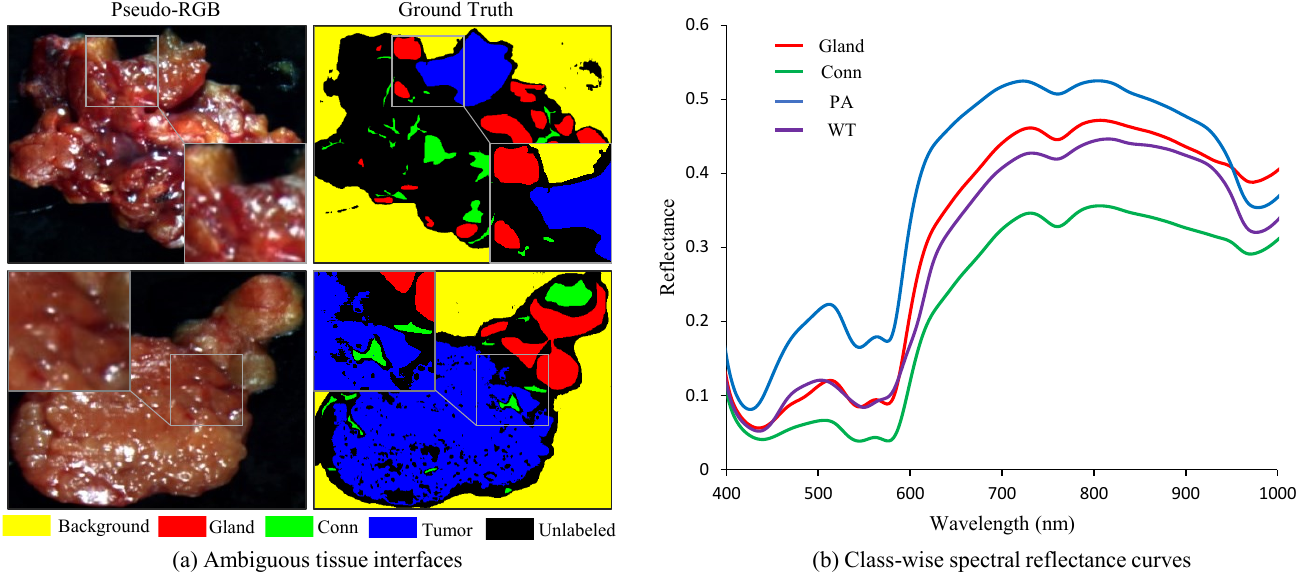}
    \caption{Spatial and spectral characteristics of SalivaryHSI-Seg.}
    \label{fig:dataset_characteristics}
\end{figure}

SalivaryHSI-Seg also exhibits task-relevant spatial and spectral characteristics. As shown in Fig.~\ref{fig:dataset_characteristics}(a), the ground-truth annotations reveal category transitions between tumors and surrounding tissues, whereas the corresponding pseudo-RGB regions provide limited boundary contrast, making these interfaces difficult to delineate from appearance alone. Meanwhile, the class-wise spectral reflectance curves in Fig.~\ref{fig:dataset_characteristics}(b) exhibit differences in both response magnitude and wavelength-dependent variation patterns across tissue categories, indicating that full-HSI data provide complementary spectral information for fine-grained tissue discrimination.

Among the 106 tumor-related samples, the tumor region accounts for 25.14\% of valid pixels on average, with a median ratio of 23.92\% and an interquartile range of 10.50\%--36.40\%. Based on tumor extent and boundary complexity, we further evaluate using two challenging subsets with small tumor and complex boundary.

%%------网络结构图--------%%
\begin{figure*}[t]
\centering
\includegraphics[width=\textwidth]{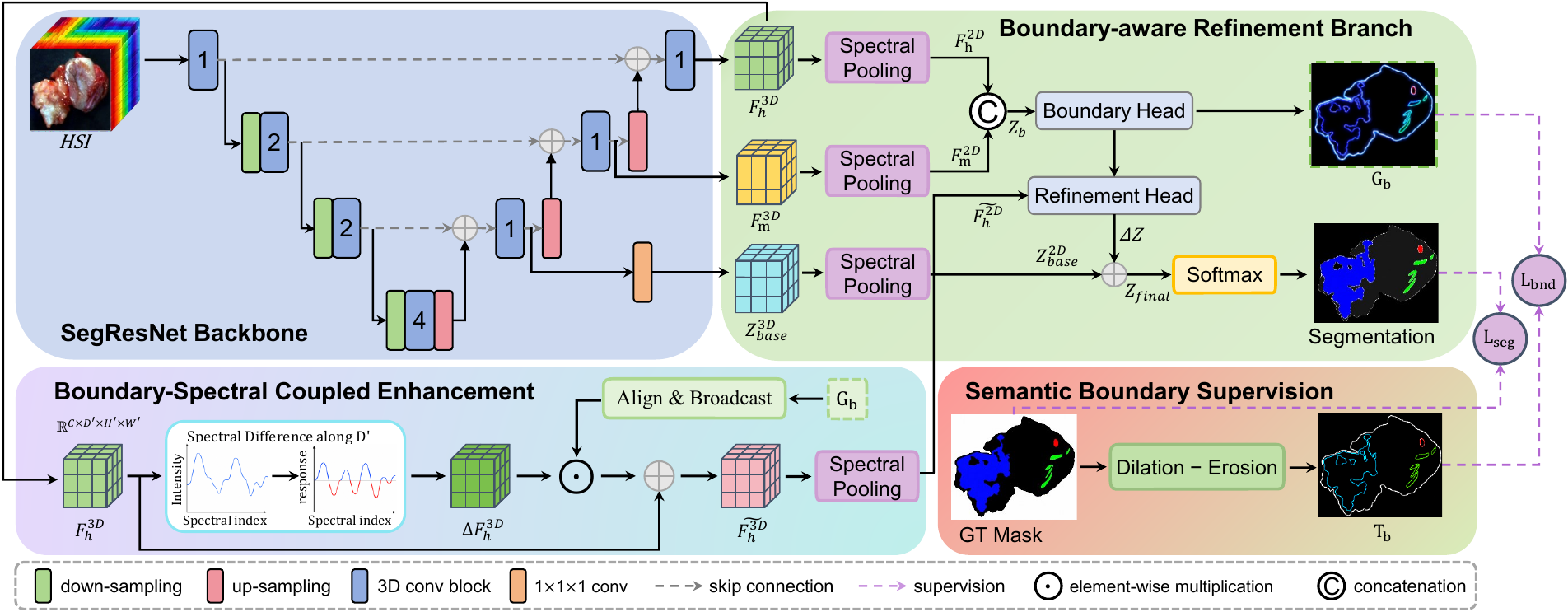}
\caption{Overview of the proposed boundary-aware full-HSI segmentation framework.}
\label{fig:framework}
\end{figure*}

% 3.3 评估协议
\subsection{Evaluation Protocol}
\label{sec:evaluation_protocol}

To evaluate cross-patient generalization, SalivaryHSI-Seg adopts a patient-level five-fold cross-validation protocol, ensuring that samples from the same patient do not appear in both the training and test sets. Across the five folds, each of the 119 samples serves as a test sample exactly once, with no patient-level leakage in any fold. Across the five folds, the training sets contain 24--26 patients with 95--96 HSI samples, while the test sets contain 5--7 patients with 23--24 HSI samples. This protocol provides a standardized setting for evaluating model generalization under a limited dataset scale.

To compare different input representations, the benchmark defines four input representations: pseudo-RGB, PCA-3, 5 bands, and full HSI. Pseudo-RGB is synthesized from three bands near 650, 550, and 450 nm to approximate conventional RGB imaging. PCA-3 is obtained by computing the first three principal components from the full HSI cube, providing a compact data-driven low-dimensional representation. The five-band representation uses five wavelengths near 450, 550, 650, 750, and 850 nm, covering representative visible and near-infrared regions while keeping the input dimensionality low. Full HSI uses all 256 continuous spectral bands. All input representations share the same annotations and data splits for fair comparison.

For evaluation, we report Dice, mean Intersection over Union (mIoU),
95th percentile Hausdorff Distance (HD95), and specificity. 
Performance is summarized at the overall and tumor-category levels,
while class-wise analysis is further provided for the full-HSI benchmark. Five-fold results are reported as mean and standard deviation. For subsets with challenging attributes, model performance is further evaluated on the small tumor and complex boundary subsets. We further apply a subtype-consistency decoding strategy to all methods: PA and WT probability masses are aggregated over valid pixels to determine the model-predicted dominant subtype.%and the non-dominant subtype is suppressed by a factor of $\exp(-\tau)$ followed by pixel-wise renormalization. We set $\tau=2.0$ for all methods without using the ground-truth diagnosis.

\section{Method}
The overall framework of HyBSD is illustrated in Fig.~\ref{fig:framework}. 
HyBSD takes the complete HSI cube as input and uses a 3D CNN backbone to extract spatial-spectral features and generates base segmentation logits. 
Subsequently, spectrally pooled features are used to predict category-related foreground boundaries under Semantic Boundary Supervision (SBS). The predicted boundary information is further utilized by Boundary-Spectral Coupled Enhancement (BSCE) to enhance spectral-difference cues in high-level 3D features, and by the Boundary-aware Refinement Branch (BAR) to generate residual logits for correcting the base prediction. The final segmentation logits are obtained by adding the residual logits to the base logits.
Next, we introduce these components in detail.

%% 输入模态总表
\begin{table*}[!t]
\caption{Comparison of different input representations using the same 2D U-Net architecture. Dice, IoU, and specificity are reported in percentage; HD95 is reported in pixels.}
\label{tab:input_representation_results}
\centering
\scriptsize
\setlength{\tabcolsep}{5pt}
\renewcommand{\arraystretch}{1.05}

\resizebox{0.55\textwidth}{!}{
\begin{tabular}{@{}llcccc@{}}
\toprule

Metric & Category
& Pseudo-RGB
& PCA-3
& 5 Bands
& Full HSI \\
\midrule

\multirow{2}{*}{Dice $\uparrow$}
& Overall
& 53.58$\pm$9.52
& 30.66$\pm$3.48
& 56.14$\pm$7.18
& \textbf{67.22$\pm$9.79} \\

& Tumor
& 60.05$\pm$12.36
& 29.82$\pm$5.34
& 64.16$\pm$9.77
& \textbf{75.56$\pm$13.22} \\

\midrule

IoU $\uparrow$
& Overall
& 43.33$\pm$9.96
& 22.93$\pm$3.26
& 45.98$\pm$8.56
& \textbf{58.99$\pm$11.21} \\

\midrule

HD95 $\downarrow$
& Overall
& 257.39$\pm$43.72
& 237.49$\pm$52.58
& 246.90$\pm$42.73
& \textbf{190.57$\pm$61.22} \\

\midrule

Specificity $\uparrow$
& Overall
& 96.25$\pm$1.76
& 94.27$\pm$1.93
& 97.00$\pm$1.31
& \textbf{97.91$\pm$1.59} \\

\bottomrule
\end{tabular}
}
\end{table*}

% 4.1 边界-光谱耦合特征增强
\subsection{Boundary-Spectral Coupled Enhancement}
%Although 3D SegResNet can directly learn spatial-spectral representations from the HSI cube, high-level semantic features may weaken fine-grained spectral variations around tissue interfaces. Since these regions often exhibit both spatial boundary ambiguity and local response variations between adjacent bands, local spectral differences can provide complementary cues for boundary-aware segmentation.

%To this end, we propose the Boundary-Spectral Coupled Enhancement module, which models first-order spectral differences in high-level 3D features and modulates the enhancement process using the Predicted Boundary generated by the Boundary Head. Given the high-level 3D feature produced by 3D SegResNet,

Although the 3D CNN backbone can learn spatial-spectral representations from the HSI cube, its high-level features are mainly optimized for region-level semantic prediction and do not explicitly emphasize local cross-band variations around tissue interfaces. Meanwhile, different salivary gland tissue categories exhibit distinct spectral reflectance patterns across wavelengths, with differences in both response magnitude and cross-band variation trends. Modeling such spectral-difference cues can therefore provide complementary information for tissue discrimination in full-HSI segmentation.

To this end, we propose Boundary-Spectral Coupled Enhancement (BSCE), which models first-order spectral differences in high-level 3D features and modulates the enhancement process using the predicted boundary information generated by the Boundary Head. Given the high-level 3D feature produced by the backbone $\mathbf{F}_{h}^{3d}$,
% \[
% \mathbf{F}_{h}^{3d} \in \mathbb{R}^{C \times D' \times H' \times W'},
% \]
% where $C$ denotes the number of channels, $D'$ denotes the spectral dimension of the feature, and $H'$ and $W'$ denote the spatial dimensions, 
we compute the first-order spectral difference along the spectral dimension:
\begin{equation}
    \Delta \mathbf{F}_{h}^{3d}
=
\mathcal{D}_{s}(\mathbf{F}_{h}^{3d}),
\end{equation}
where $\mathcal{D}_{s}(\cdot)$ denotes the spectral-difference operator. The resulting $\Delta \mathbf{F}_{h}^{3d}$ corresponds to the Spectral Cue in Figure~\ref{fig:framework}.

However, directly enhancing spectral differences at all positions may introduce irrelevant responses from background, noise, or tissue interiors. Therefore, we derive a boundary gating map $\mathbf{G}_{b}$ from the Predicted Boundary and use it as a spatial modulation signal. After aligning $\mathbf{G}_{b}$ to the spatial resolution of $\mathbf{F}_{h}^{3d}$ and broadcasting it along the spectral dimension, the enhanced high-level feature is defined as
\begin{equation}
\hat{\mathbf{F}}_{h}^{3d}
=
\mathbf{F}_{h}^{3d}
+
\gamma
\left(
\alpha_0 + \alpha_1 \mathbf{G}_{b}
\right)
\odot
\Delta \mathbf{F}_{h}^{3d},
\end{equation}
where $\odot$ denotes element-wise multiplication, $\gamma$ is a learnable scaling factor, and $\alpha_0$ and $\alpha_1$ control the base spectral enhancement strength and the boundary-guided enhancement strength, respectively.
This residual enhancement form preserves the original high-level spatial-spectral representation while selectively injecting boundary-sensitive spectral-difference information. The enhanced feature $\hat{\mathbf{F}}_{h}^{3d}$ is then transformed into $\hat{\mathbf{F}}_{h}^{2d}$ through Spectral Pooling and fed into the Boundary-aware Refinement Branch for final segmentation refinement.

%  4.2 语义边界感知细化
\subsection{Semantic Boundary-Aware Refinement}
In addition to spectral-difference cues, we observe from the SalivaryHSI-Seg that foreground tissues such as glandular parenchyma, connective tissue septa, PA tumor, and WT tumor may be spatially adjacent to each other. If only a single boundary map is used for supervision, information from different tissue interfaces may be mixed, making it difficult to provide fine-grained structural constraints for multi-class segmentation. To this end, we design a Boundary-aware Refinement Branch and introduce Semantic Boundary Supervision to model foreground category-related boundary responses and feed them back as structural cues for final segmentation prediction.

Specifically, Spectral Pooling is first applied to the backbone outputs $\mathbf{F}_{h}^{3d}$, $\mathbf{F}_{m}^{3d}$, and $\mathbf{Z}_{base}^{3d}$ to obtain the high-level feature $\mathbf{F}_{h}^{2d}$, the intermediate feature $\mathbf{F}_{m}^{2d}$, and the base logits $\mathbf{Z}_{base}^{2d}$, respectively. The high-level feature provides semantic information, while the intermediate feature complements local structural details. These two features are concatenated and fed into the Boundary Head to produce semantic boundary logits:
\begin{equation}
\mathbf{Z}_{b}
=
\psi_{b}
\left(
[
\mathbf{F}_{h}^{2d},
\mathbf{F}_{m}^{2d}
]
\right),
\quad
% \mathbf{Z}_{b}
% \in
% \mathbb{R}^{(K-1) \times H \times W}, 
\end{equation}
% where $K$ denotes the number of segmentation classes, and the $K-1$ boundary channels correspond to foreground tissue categories excluding the background. 
During training, the Predicted Boundary $\sigma(\mathbf{Z}_{b})$ is supervised by the $\mathbf{T}_{b}$  generated from the segmentation mask.

To feed boundary information back into segmentation prediction, the Refinement Head uses the enhanced feature $\hat{\mathbf{F}}_{h}^{2d}$ produced by BSCE and the Predicted Boundary to predict residual logits:
\begin{equation}
\mathbf{Z}_{final}
=
\mathbf{Z}_{base}^{2d}
+
\Delta \mathbf{Z},
\quad
\Delta \mathbf{Z}
=
\phi_{r}
\left(
[
\hat{\mathbf{F}}_{h}^{2d},
\sigma(\mathbf{Z}_{b})
]
\right),
\end{equation}
where $\Delta \mathbf{Z}$ denotes the boundary-aware residual correction term. In this way, the 3D backbone provides the base tissue-region prediction, while the Boundary-aware Refinement Branch uses the enhanced high-level feature and semantic boundary information for local refinement. The final segmentation map is obtained by applying Softmax to $\mathbf{Z}_{final}$.

%% 合并总表+分类别表
\begin{table*}[!t]
\caption{Full-HSI benchmark results. Dice, IoU, and specificity are reported in percentage; HD95 is reported in pixels.}
\label{tab:full_hsi_results}
\centering
\scriptsize
\setlength{\tabcolsep}{3.2pt}
\renewcommand{\arraystretch}{1.05}

\resizebox{\textwidth}{!}{
\begin{tabular}{@{}llcccccccc@{}}
\toprule
& &
\multicolumn{1}{c}{\textit{2D CNN}} &
\multicolumn{4}{c}{\textit{3D CNN}} &
\multicolumn{3}{c}{\textit{3D Transformer}} \\
\cmidrule(lr){3-3}
\cmidrule(lr){4-7}
\cmidrule(lr){8-10}

Metric & Category
& 2D U-Net
& 3D U-Net
& VNet
& SegResNet
& HyBSD (Ours)
& UNETR
& SwinUNETR
& HyBSD-Swin \\
\midrule

% ================= Dice =================
\multirow{6}{*}{Dice $\uparrow$}
& \textbf{Overall}
& 67.22$\pm$9.79
& 62.61$\pm$9.08
& 46.97$\pm$4.90
& 60.21$\pm$6.84
& \textbf{74.97$\pm$7.90}
& 38.43$\pm$13.23
& 56.83$\pm$13.73
& 66.87$\pm$10.87 \\

& Gland
& 59.05$\pm$11.48
& 46.37$\pm$10.51
& 25.83$\pm$7.03
& 43.50$\pm$5.40
& \textbf{61.65$\pm$11.71}
& 29.94$\pm$7.04
& 42.94$\pm$11.77
& 55.93$\pm$12.61 \\

& Conn
& 61.98$\pm$9.33
& 53.11$\pm$8.86
& 43.03$\pm$4.58
& 50.04$\pm$4.16
& \textbf{62.71$\pm$7.25}
& 31.36$\pm$13.15
& 42.45$\pm$19.23
& 59.32$\pm$4.22 \\

& PA
& 74.87$\pm$16.23
& 78.34$\pm$12.10
& 66.43$\pm$22.74
& 65.04$\pm$28.04
& \textbf{84.19$\pm$10.75}
& 60.05$\pm$11.16
& 62.65$\pm$20.11
& 78.31$\pm$12.27 \\

& WT
& 89.76$\pm$8.68
& 72.63$\pm$28.59
& 52.59$\pm$30.24
& 82.27$\pm$14.59
& \textbf{91.33$\pm$8.88}
& 32.38$\pm$44.70
& 79.29$\pm$16.28
& 73.91$\pm$31.07 \\

& Tumor
& 75.56$\pm$13.22
& 75.48$\pm$14.88
& 59.51$\pm$11.01
& 73.66$\pm$12.50
& \textbf{87.76$\pm$7.08}
& 46.21$\pm$20.75
& 70.97$\pm$16.43
& 76.11$\pm$15.83 \\

\midrule

% ================= IoU =================
\multirow{5}{*}{IoU $\uparrow$}
& \textbf{Overall}
& 58.99$\pm$11.21
& 53.28$\pm$10.13
& 39.59$\pm$4.45
& 51.76$\pm$6.56
& \textbf{67.55$\pm$9.30}
& 30.43$\pm$12.15
& 47.05$\pm$13.99
& 57.96$\pm$11.66 \\

& Gland
& 50.54$\pm$13.23
& 36.27$\pm$8.71
& 20.57$\pm$6.80
& 34.11$\pm$4.61
& \textbf{53.37$\pm$12.09}
& 21.03$\pm$6.08
& 32.41$\pm$10.57
& 46.56$\pm$11.76 \\

& Conn
& 49.60$\pm$9.49
& 40.56$\pm$8.83
& 31.66$\pm$4.58
& 37.19$\pm$3.91
& \textbf{49.84$\pm$7.51}
& 21.16$\pm$10.24
& 31.28$\pm$15.98
& 45.85$\pm$4.37 \\

& PA
& 69.38$\pm$17.88
& 70.48$\pm$14.41
& 59.77$\pm$23.76
& 58.15$\pm$26.32
& \textbf{78.77$\pm$13.38}
& 50.70$\pm$10.64
& 53.75$\pm$23.33
& 71.18$\pm$15.06 \\

& WT
& 84.64$\pm$12.43
& 65.81$\pm$30.99
& 46.36$\pm$26.47
& 77.59$\pm$15.30
& \textbf{88.23$\pm$10.37}
& 28.82$\pm$39.88
& 70.76$\pm$19.98
& 68.26$\pm$32.71 \\

\midrule

% ================= HD95 =================
\multirow{5}{*}{HD95 $\downarrow$}
& \textbf{Overall}
& 190.57$\pm$61.22
& 187.24$\pm$35.45
& 163.74$\pm$27.08
& 190.03$\pm$40.24
& \textbf{133.96$\pm$37.01}
& 219.27$\pm$21.97
& 222.65$\pm$59.65
& 191.91$\pm$31.24 \\

& Gland
& 256.05$\pm$79.02
& 292.15$\pm$44.17
& 264.25$\pm$71.96
& 241.52$\pm$40.11
& \textbf{190.16$\pm$78.88}
& 323.18$\pm$37.91
& 284.67$\pm$76.35
& 266.54$\pm$71.16 \\

& Conn
& 222.46$\pm$27.51
& 226.31$\pm$22.48
& 230.45$\pm$26.67
& 215.69$\pm$19.43
& \textbf{205.63$\pm$26.91}
& 264.66$\pm$32.42
& 240.74$\pm$11.49
& 214.91$\pm$26.13 \\

& PA
& 158.14$\pm$106.98
& 139.73$\pm$69.99
& 119.09$\pm$43.50
& 174.26$\pm$104.62
& \textbf{109.72$\pm$62.50}
& 196.04$\pm$103.08
& 180.22$\pm$150.04
& 165.56$\pm$51.12 \\

& WT
& 123.62$\pm$121.91
& 90.78$\pm$62.93
& 41.17$\pm$35.87
& 128.65$\pm$68.64
& \textbf{30.34$\pm$16.93}
& 93.19$\pm$127.67
& 184.95$\pm$90.16
& 120.65$\pm$102.65 \\

\midrule

% ================= Specificity =================
\multirow{5}{*}{Specificity $\uparrow$}
& \textbf{Overall}
& 97.91$\pm$1.59
& 97.58$\pm$1.14
& 95.95$\pm$1.82
& 97.20$\pm$1.65
& \textbf{98.65$\pm$1.14}
& 95.14$\pm$1.54
& 96.93$\pm$1.75
& 97.96$\pm$1.52 \\

& Gland
& 98.42$\pm$0.65
& 97.34$\pm$1.95
& 93.75$\pm$7.02
& 98.14$\pm$2.02
& \textbf{98.45$\pm$1.71}
& 94.30$\pm$4.06
& 95.63$\pm$2.71
& 97.39$\pm$2.00 \\

& Conn
& 97.21$\pm$2.12
& 95.92$\pm$1.97
& 93.34$\pm$2.75
& 96.63$\pm$1.85
& \textbf{97.62$\pm$1.55}
& 93.46$\pm$6.00
& 96.75$\pm$1.20
& 96.86$\pm$2.09 \\

& PA
& 98.64$\pm$1.56
& 98.03$\pm$2.74
& 98.26$\pm$1.42
& 97.56$\pm$2.94
& \textbf{98.79$\pm$1.32}
& 95.14$\pm$4.18
& 97.16$\pm$2.93
& 97.74$\pm$2.35 \\

& WT
& 98.22$\pm$2.48
& 99.03$\pm$1.02
& 98.45$\pm$3.42
& 96.48$\pm$5.53
& 99.73$\pm$0.28
& 97.66$\pm$4.46
& 98.20$\pm$1.73
& \textbf{99.84$\pm$0.10} \\

\bottomrule
\end{tabular}
}
\end{table*}

% 4.3 训练目标
\subsection{Training Objective}
The training of HyBSD is supervised by the segmentation and boundary-aware branches. 
For segmentation branch, the loss is computed using the final logits $\mathbf{Z}_{final}$ and consists of a negative log-likelihood loss and a Dice loss:
\begin{equation}
\mathcal{L}_{seg}
=
\mathcal{L}_{nll}
+
\mathcal{L}_{dice}. 
\end{equation}
Unlabeled regions are excluded from loss computation, and the Dice term is used to mitigate the effect of class imbalance on foreground tissue learning.
For semantic boundary supervision, category-wise boundary targets are generated from the GT Mask by applying morphological operations to each foreground tissue category. Boundary supervision is applied only to foreground categories excluding the background, and unlabeled regions are excluded from the boundary loss. The boundary loss consists of a binary cross-entropy loss and a boundary Dice loss:
\begin{equation}
\mathcal{L}_{bnd}
=
\mathcal{L}_{bce}
+
\mathcal{L}_{bdice}.
\end{equation}

The overall loss is defined as
\begin{equation}
\mathcal{L}
=
\mathcal{L}_{seg}
+
\lambda_{b}\mathcal{L}_{bnd},    
\end{equation}
where $\mathcal{L}_{seg}$ denotes the segmentation loss, $\mathcal{L}_{bnd}$ denotes the boundary loss, and $\lambda_{b}$ is the boundary loss weight.

\section{Experiments}
% 5.1 实验设置
\subsection{Experimental Settings}
\label{sec:experimental_settings}

All experiments follow the official evaluation protocol described in Section~\ref{sec:evaluation_protocol}. We evaluate the four predefined input representations using the same 2D U-Net architecture, and compare representative full-HSI segmentation models, including 2D U-Net~\cite{ronneberger2015u}, 3D U-Net~\cite{cciccek20163d}, VNet~\cite{milletari2016v}, SegResNet~\cite{myronenko20183d}, UNETR~\cite{hatamizadeh2022unetr}, SwinUNETR~\cite{tang2022self}, and our method. All ablation variants use the same data splits.

During training, spatial patches of size $128\times128$ are sampled
from each image. The batch size is set to 1, and all models are trained
for 30 epochs. We use Adam with an initial learning rate of $3\times10^{-4}$, a weight decay of $5\times10^{-4}$, and a cosine warm-restart scheduler. The random seed is set to 42. Standard baselines are trained with a cross-entropy Dice loss, while our boundary-aware model uses the joint segmentation and boundary supervision objective in Section~IV-C with $\lambda_b=0.1$.

During inference, we perform full-image tiled prediction with a tile target size of $128 \times 128$ and a contextual margin of 16 pixels. %Since each tumor-related sample contains a single diagnosis-specific tumor subtype, we apply a unified subtype-consistency decoding protocol to reduce subtype-level false positives. %PA and WT probability masses are computed from model predictions over valid pixels, and predictions of the weaker subtype are suppressed when the dominant subtype exceeds a fixed threshold $\tau=2.0$; the same $\tau$ is used for all methods, without using the ground-truth diagnosis.

% 5.2 模态分析
\subsection{Results for Different Input Bands}

Table~\ref{tab:input_representation_results} compares the four input representations using the same 2D U-Net architecture. Full HSI achieves the best performance among the four input representations, including pseudo-RGB, PCA-3, 5 bands, and full HSI, indicating that the complete spectral cube preserves useful tissue-discriminative information. PCA-3 performs the worst, suggesting that aggressive spectral compression may discard important cues. Although 5 bands outperform pseudo-RGB, their gap to full HSI shows that a small number of bands is insufficient to fully characterize spectral differences among tissue categories.
%%添加
In particular, compared with the 5-band representation, full HSI improves the overall Dice and tumor Dice by 11.08 and 11.40 percentage points, respectively, while reducing HD95 by 56.33 pixels. These results further motivate the use of full HSI as the default input representation in the subsequent benchmark experiments.

\subsection{Overall Performance}

Table~\ref{tab:full_hsi_results} further compares representative full-HSI segmentation models, including 2D U-Net, 3D CNNs, Transformer-based models, and our method. Our method achieves the best performance on all main metrics, with 74.97\% mDice, 67.55\% mIoU, 133.96 HD95, 98.65\% mSpec, and 87.76\% Tumor Dice. Compared with 3D CNN backbone, it improves mDice by 14.76 percentage points, mIoU by 15.79 percentage points, and Tumor Dice by 14.10 percentage points, while reducing HD95 by 56.07 pixels. These results show the effectiveness of boundary-aware spatial-spectral modeling for full-HSI tissue segmentation.

Under the current setting, Transformer-based baselines perform worse than CNN-based full-HSI models, suggesting that directly transferring existing 3D Transformer segmentation models to medical HSI still requires task-specific adaptation. Table~\ref{tab:full_hsi_results} further reports category-specific results. %Specificity is included as a supplementary metric and mainly interpreted under the full-HSI setting to avoid over-emphasizing conservative predictions under class imbalance. 
Our method achieves the best Dice and IoU across all foreground categories, indicating that the improvement extends to both tumor and non-tumor tissue structures.

To further examine the generalization of the proposed design to different architectures, we incorporate the proposed components into SwinUNETR and denote the resulting variant as HyBSD-Swin. As shown in Table~\ref{tab:full_hsi_results}, HyBSD-Swin achieves improved overall segmentation performance compared with the original SwinUNETR. This result indicates that the proposed boundary-aware spatial-spectral modeling strategy can also benefit Transformer-based architectures.

% 5.3 挑战属性分析
\subsection{Results for Challenging Subsets}

To further evaluate difficult tumor-related cases, we analyze two subsets with challenging attributes: small tumor and complex boundary. For each PA/WT sample, the tumor area ratio $r_t$ and boundary complexity score $s_b$ are defined as
\begin{equation}
r_t = \frac{|\Omega_t|}{|\Omega_f|},
\quad
s_b = \frac{|\partial \Omega_t|}{\sqrt{|\Omega_t|}}.
\label{eq:challenge_metrics}
\end{equation}
where $\Omega_t$ is the diagnosis-specific tumor region, $\Omega_f$ is the valid foreground tissue region, and $\partial \Omega_t$ is the tumor boundary. Samples in the lower tertile of $r_t$ form the small tumor subset, while samples in the upper tertile of $s_b$ form the complex boundary subset. Tumor-related metrics are computed using PA-class metrics for PA samples and WT-class metrics for WT samples.

%%-----挑战属性小表----%%
\begin{table}[t]
\caption{Results of challenging subsets.}
\label{tab:challenge_attribute}
\centering
\scriptsize
\begin{tabular}{lccc}
\toprule
\multicolumn{4}{c}{\textbf{Small tumor: 36 (15/21)}} \\
\midrule
Method & Tumor Dice $\uparrow$ & Tumor HD95 $\downarrow$ & mDice$_p$ $\uparrow$ \\
\midrule
Full-HSI U-Net & 61.46$\pm$38.07 & 360.01$\pm$408.22 & 69.20$\pm$18.76 \\
SegResNet & 60.16$\pm$38.70 & 360.19$\pm$427.20 & 61.51$\pm$21.87 \\
HyBSD (Ours)  & \textbf{76.29$\pm$35.71} & \textbf{166.09$\pm$335.00} & \textbf{75.07$\pm$21.20} \\
\midrule
\multicolumn{4}{c}{\textbf{Complex boundary: 36 (24/12)}} \\
\midrule
Method & Tumor Dice $\uparrow$ & Tumor HD95 $\downarrow$ & mDice$_p$ $\uparrow$ \\
\midrule
Full-HSI U-Net & 71.99$\pm$38.07 & 245.61$\pm$435.23 & 58.21$\pm$24.16 \\
SegResNet & 72.31$\pm$37.64 & 197.77$\pm$368.10 & 50.25$\pm$19.84 \\
HyBSD (Ours)  & \textbf{91.01$\pm$19.61} & \textbf{42.63$\pm$146.50} & \textbf{70.00$\pm$21.97} \\
\bottomrule
\end{tabular}
\end{table}

\begin{figure*}[!t]
\centering
\includegraphics[width=\textwidth]{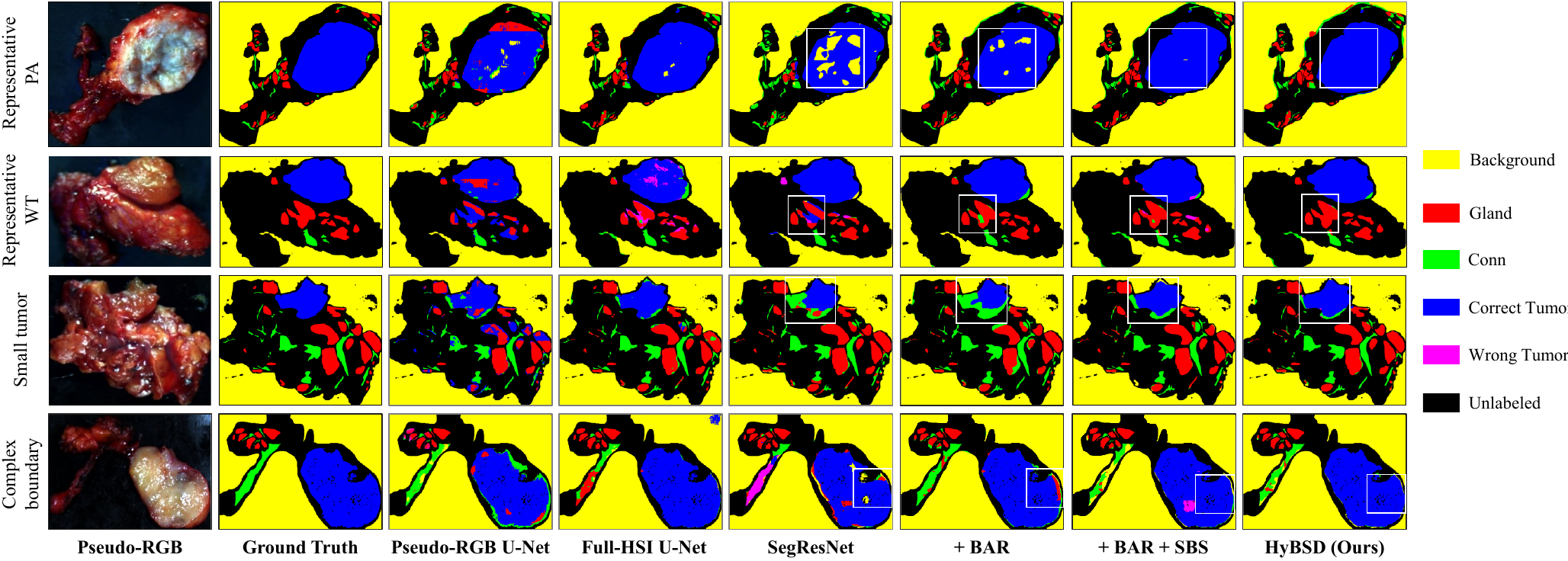}
\caption{Qualitative segmentation results on representative and challenging samples. Correct Tumor denotes the diagnosis-consistent tumor class, Wrong Tumor denotes the other tumor class, and Unlabeled denotes unannotated regions.}
\label{fig:qualitative}
\end{figure*}

Table~\ref{tab:challenge_attribute} reports the results on the two challenge subsets. Our method consistently achieves higher Tumor Dice and mDice$_p$ with lower Tumor HD95 than Full-HSI U-Net and 3D CNN backbone. These results, together with Fig.~\ref{fig:qualitative}, show that the challenge subsets reveal performance differences beyond overall average evaluation.

%%----可视化分割图-------%%%

% 5.4 消融实验
\subsection{Ablation Study}

To validate the effectiveness of the proposed components, we conduct ablation experiments on both CNN and Transformer backbones by progressively adding boundary-aware refinement, semantic boundary supervision, and boundary-spectral coupled enhancement. BAR, SBS, and BSCE denote these three components, respectively. As shown in Table~\ref{tab:ablation}, the first row corresponds to the backbone baseline, while the last row represents the full proposed method.

%%-----消融实验表格-----%%
\begin{table}[!t]
\caption{Ablation study of BAR, SBS, and BSCE on CNN and Transformer backbones.}
\label{tab:ablation}
\centering
\scriptsize
\setlength{\tabcolsep}{3pt}
\renewcommand{\arraystretch}{1.05}

\resizebox{\columnwidth}{!}{%
\begin{tabular}{ccc|cccc}
\toprule
\multicolumn{3}{c|}{Components}
&
\multicolumn{4}{c}{Performance} \\
\cmidrule(lr){1-3}
\cmidrule(lr){4-7}

BAR & SBS & BSCE
& mDice $\uparrow$
& mIoU $\uparrow$
& HD95 $\downarrow$
& Tumor Dice $\uparrow$ \\
\midrule

\multicolumn{7}{c}{\textit{SegResNet}} \\
\midrule

-- & -- & --
& 60.21$\pm$6.84
& 51.76$\pm$6.56
& 190.03$\pm$40.24
& 73.66$\pm$12.50 \\

$\surd$ & -- & --
& 72.12$\pm$9.09
& 63.99$\pm$10.53
& 166.50$\pm$38.70
& 84.34$\pm$8.84 \\

$\surd$ & $\surd$ & --
& 74.13$\pm$8.33
& 66.86$\pm$9.34
& 152.87$\pm$51.31
& 86.54$\pm$10.15 \\

$\surd$ & $\surd$ & $\surd$
& \textbf{74.97$\pm$7.90}
& \textbf{67.55$\pm$9.30}
& \textbf{133.96$\pm$37.01}
& \textbf{87.76$\pm$7.08} \\

\midrule

\multicolumn{7}{c}{\textit{SwinUNETR}} \\
\midrule

-- & -- & --
& 56.83$\pm$13.73
& 47.05$\pm$13.99
& 222.65$\pm$59.65
& 70.97$\pm$16.43 \\

$\surd$ & $\surd$ & --
& 65.11$\pm$7.89
& 56.75$\pm$9.32
& 210.14$\pm$50.88
& 73.01$\pm$11.05 \\

$\surd$ & $\surd$ & $\surd$
& \textbf{66.87$\pm$10.87}
& \textbf{57.96$\pm$11.66}
& \textbf{191.91$\pm$31.24}
& \textbf{76.11$\pm$15.83} \\

\bottomrule
\end{tabular}%
}

\end{table}

Compared with the backbone baseline, adding BAR improves mDice from 60.21\% to 72.12\% and Tumor Dice from 73.66\% to 84.34\%, while reducing HD95 from 190.03 to 166.50. This indicates that boundary-aware refinement provides useful structural cues for the 3D backbone. Further introducing SBS improves mDice, mIoU, and Tumor Dice to 74.13\%, 66.86\%, and 86.54\%, respectively, suggesting that semantic boundary supervision provides finer category-related boundary constraints for multi-class tissue segmentation.

Adding BSCE further improves all metrics, increasing mDice and Tumor Dice to 74.97\% and 87.76\%, while reducing HD95 to 133.96. These results show that BAR and SBS mainly improve region segmentation, while BSCE further enhances boundary localization and leads to the best overall performance.

%%-----HyBSD component visualization-----%%
\begin{figure}[!t]
\centering
\includegraphics[width=\columnwidth]
{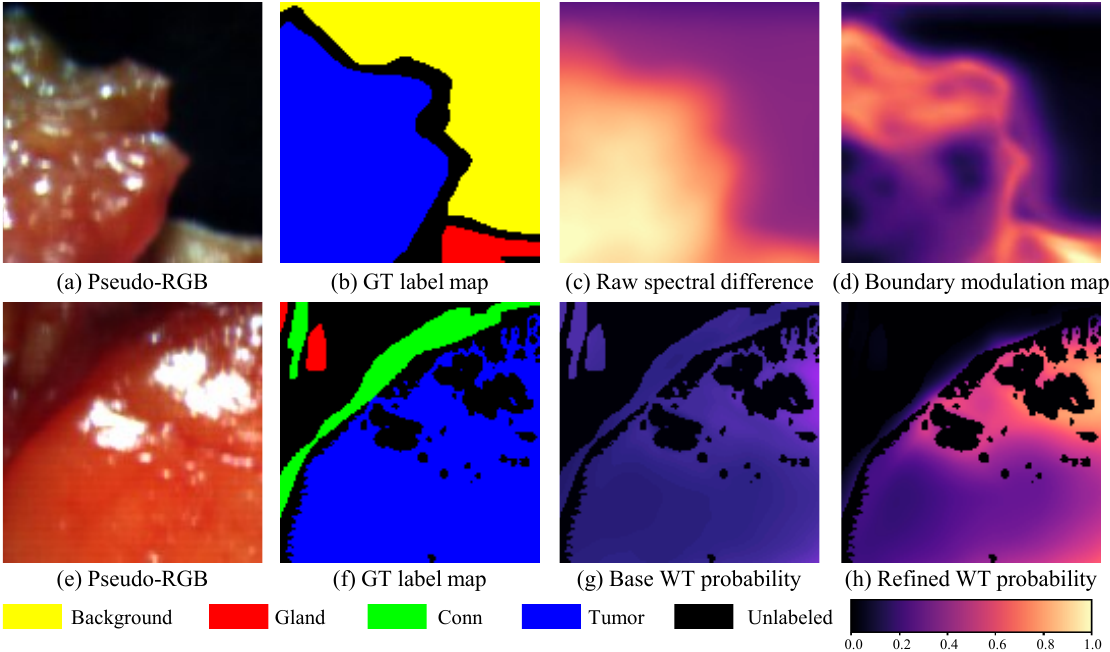}
\caption{Intermediate-response visualization of HyBSD. Top: BSCE,
including the pseudo-RGB image, ground truth, normalized raw
spectral-difference response, and normalized boundary modulation
weight. Bottom: boundary-aware refinement, including the pseudo-RGB
image, ground truth, and WT-class probability maps before and after
refinement.}
\label{fig:component_visualization}
\end{figure}

To further investigate whether these improvements are dependent on a specific backbone, we evaluate the proposed components on the SwinUNETR architecture. As shown in Table~\ref{tab:ablation}, progressively incorporating SBS, BAR, and BSCE improves mDice from 56.83\% to 66.87\% and mIoU from 47.05\% to 57.96\%, while reducing HD95 from 222.65 to 191.91 pixels. These improvements demonstrate that the proposed components can be effectively extended to Transformer-based architectures.

As highlighted by the white boxes in Fig.~\ref{fig:qualitative}, 
BAR improves local region coherence, SBS further reduces boundary-related 
misclassification, and the full HyBSD produces cleaner predictions around 
ambiguous tissue interfaces. Fig.~\ref{fig:component_visualization} further 
visualizes the corresponding intermediate responses. The raw first-order 
spectral-difference map captures local cross-band variations, while boundary 
modulation emphasizes tissue interfaces. After refinement, the WT-class 
probability becomes more spatially coherent and better aligned with the 
annotated region. These observations are consistent with the quantitative 
improvements in Table~\ref{tab:ablation}.

\section{Conclusion}

In this paper, we introduced SalivaryHSI-Seg, a hyperspectral tissue segmentation benchmark for ex vivo salivary gland specimens, containing 119 HSI samples from 31 patients with multi-class pixel-level annotations. We established a patient-level five-fold evaluation protocol. We further proposed HyBSD, a hyperspectral boundary-guided spectral-difference network for salivary gland tissue segmentation, which exploits boundary-guided spectral-difference cues to improve tissue delineation in ambiguous transition regions. Extensive experiments show that full-HSI input provides stronger tissue-discriminative information, and that HyBSD achieves better performance on overall metrics and difficult tumor-related samples. We hope SalivaryHSI-Seg will support future research on medical HSI tissue segmentation.

% \section*{Acknowledgment}

% This work was supported in part by the National Natural Science Foundation of China under Grant 62576216 and the Guangdong Provincial Key Laboratory under Grant 2023B1212060076. It was also supported by the Yunnan Provincial Clinical Medical Center Research Project under Grants 2024YNLCYXZX0227 and 2024YNLCYXZX0229, the Yunnan Clinical Research Center for Oral Diseases under Grant 202505AJ310001, and the Kunming Medical University Education and Teaching Research Project under Grant 2026-JY-Z-18.

% \section*{Acknowledgment}
% We express our gratitude to the codes of PatchDiffusion-Pytorch~\cite{luhman2022improving} and guided-diffusion~\cite{dhariwal2021diffusion} that facilitate the pre-training of our diffusion model. 
% We also thank the codes of helix2fan~\cite{wagner2022dual} and torch-radon~\cite{torch_radon} that enable us to validate the effectiveness of our method on clinical data. We have made our codes and pre-trained models publicly available.

\section*{Acknowledgments}
This work was supported in part by the National Natural Science Foundation of China under Grant 62576216 and the Guangdong Provincial Key Laboratory under Grant 2023B1212060076. This work was also supported by the Yunnan Provincial Clinical Medical Center Research Project under Grants 2024YNLCYXZX0227 and 2024YNLCYXZX0229, the Yunnan Clinical Research Center for Oral Diseases under Grant 202505AJ310001, and the Kunming Medical University Education and Teaching Research Project under Grant 2026-JY-Z-18.

\bibliographystyle{ieee_fullname}
\bibliography{tmi2023}

@article{lu2014medical,
  title={Medical hyperspectral imaging: a review},
  author={Lu, Guolan and Fei, Baowei},
  journal={Journal of biomedical optics},
  volume={19},
  number={1},
  pages={010901--010901},
  year={2014},
  publisher={Society of Photo-Optical Instrumentation Engineers}
}

@article{tran2026medical,
  title={Medical hyperspectral imaging: an updated review of technology advancements and biomedical applications},
  author={Tran, Minh Ha and Ma, Ling and Yuan, Mandy and Fei, Baowei},
  journal={Journal of Biomedical Optics},
  volume={31},
  number={03},
  pages={30901},
  year={2026},
  publisher={SPIE-International Society for}
}

@article{ali2025surgical,
  title={Surgical hyperspectral imaging: A systematic review},
  author={Ali, Hafsa Moontari and Xiao, Yiming and Kersten-Oertel, Marta},
  journal={Computer Assisted Surgery},
  volume={30},
  number={1},
  pages={2546819},
  year={2025},
  publisher={Taylor \& Francis}
}

@article{thiem2021hyperspectral,
  title={Hyperspectral imaging and artificial intelligence to detect oral malignancy--part 1-automated tissue classification of oral muscle, fat and mucosa using a light-weight 6-layer deep neural network},
  author={Thiem, Daniel GE and R{\"o}mer, Paul and Gielisch, Matthias and Al-Nawas, Bilal and Schl{\"u}ter, Martin and Pla{\ss}, Bastian and K{\"a}mmerer, Peer W},
  journal={Head \& face medicine},
  volume={17},
  number={1},
  pages={38},
  year={2021},
  publisher={Springer}
}

@article{roddan2024calibration,
  title={Calibration-Jitter: Augmentation of hyperspectral data for improved surgical scene segmentation},
  author={Roddan, Alfie and Czempiel, Tobias and Elson, Daniel S and Giannarou, Stamatia},
  journal={Healthcare Technology Letters},
  volume={11},
  number={6},
  pages={345--354},
  year={2024},
  publisher={Wiley Online Library}
}

@article{trajanovski2020tongue,
  title={Tongue tumor detection in hyperspectral images using deep learning semantic segmentation},
  author={Trajanovski, Stojan and Shan, Caifeng and Weijtmans, Pim JC and de Koning, Susan G Brouwer and Ruers, Theo JM},
  journal={IEEE transactions on biomedical engineering},
  volume={68},
  number={4},
  pages={1330--1340},
  year={2020},
  publisher={IEEE}
}

@article{zhang2021explainable,
  title={Explainable liver tumor delineation in surgical specimens using hyperspectral imaging and deep learning},
  author={Zhang, Yating and Yu, Si and Zhu, Xueyu and Ning, Xuefei and Liu, Wei and Wang, Chuting and Liu, Xiaohu and Zhao, Ding and Zheng, Yongchang and Bao, Jie},
  journal={Biomedical optics express},
  volume={12},
  number={7},
  pages={4510--4529},
  year={2021},
  publisher={Optical Society of America}
}

@inproceedings{roddan2025samsa,
  title={SAMSA: Segment Anything Model Enhanced with Spectral Angles for Hyperspectral Interactive Medical Image Segmentation},
  author={Roddan, Alfie and Czempiel, Tobias and Xu, Chi and Elson, Daniel S and Giannarou, Stamatia},
  booktitle={International Conference on Medical Image Computing and Computer-Assisted Intervention},
  pages={478--488},
  year={2025},
  organization={Springer}
}

@article{10946212,
  title={Boundary-guided contrastive learning for semi-supervised medical image segmentation},
  author={Yang, Yang and Zhuang, Jiaxin and Sun, Guoying and Wang, Ruixuan and Su, Jingyong},
  journal={IEEE Transactions on Medical Imaging},
  volume={44},
  number={7},
  pages={2973--2988},
  year={2025},
  publisher={IEEE}
}

@article{10723767,
  title={Effective semi-supervised medical image segmentation with probabilistic representations and prototype learning},
  author={Yuan, Yuchen and Wang, Xi and Yang, Xikai and Heng, Pheng-Ann},
  journal={IEEE Transactions on Medical Imaging},
  volume={44},
  number={3},
  pages={1181--1193},
  year={2024},
  publisher={IEEE}
}

@inproceedings{KERVADEC2021101851,
  title={Boundary loss for highly unbalanced segmentation},
  author={Kervadec, Hoel and Bouchtiba, Jihene and Desrosiers, Christian and Granger, Eric and Dolz, Jose and Ayed, Ismail Ben},
  booktitle={International conference on medical imaging with deep learning},
  pages={285--296},
  year={2019},
  organization={PMLR}
}

@article{qin2025semi,
  title={Semi-supervised medical hyperspectral image segmentation using adversarial consistency constraint learning and cross indication network},
  author={Qin, Geng and Liu, Huan and Zhang, Xueyu and Li, Wei and Guo, Yuxing and Guo, Chuanbin},
  journal={IEEE Transactions on Image Processing},
  year={2025},
  publisher={IEEE}
}

@article{cao2025boundary,
  title={Boundary aware microscopic hyperspectral pathology image segmentation network guided by information entropy weight},
  author={Cao, Xueying and Gao, Hongmin and Qin, Ting and Zhu, Min and Zhang, Ping and Xu, Peipei},
  journal={Frontiers in Oncology},
  volume={15},
  pages={1549544},
  year={2025},
  publisher={Frontiers}
}

@ARTICLE{10381748,
  author={Zhang, Xueyu and Li, Qingxiang and Li, Wei and Guo, Yuxing and Zhang, Jianyun and Guo, Chuanbin and Chang, Kan and Lovell, Nigel H.},
  journal={IEEE Journal of Biomedical and Health Informatics}, 
  title={FD-Net: Feature Distillation Network for Oral Squamous Cell Carcinoma Lymph Node Segmentation in Hyperspectral Imagery}, 
  year={2024},
  volume={28},
  number={3},
  pages={1552-1563},
}

@article{yun2023spectr,
  title={SpecTr: Spectral transformer for microscopic hyperspectral pathology image segmentation},
  author={Yun, Boxiang and Lei, Baiying and Chen, Jieneng and Wang, Huiyu and Qiu, Song and Shen, Wei and Li, Qingli and Wang, Yan},
  journal={IEEE Transactions on Circuits and Systems for Video Technology},
  volume={34},
  number={6},
  pages={4610--4624},
  year={2023},
  publisher={IEEE}
}

@article{leon2023hyperspectral,
  title={Hyperspectral imaging benchmark based on machine learning for intraoperative brain tumour detection},
  author={Leon, Raquel and Fabelo, Himar and Ortega, Samuel and Cruz-Guerrero, Ines A and Campos-Delgado, Daniel Ulises and Szolna, Adam and Pi{\~n}eiro, Juan F and Espino, Carlos and O’Shanahan, Aruma J and Hernandez, Maria and others},
  journal={npj Precision Oncology},
  volume={7},
  number={1},
  pages={119},
  year={2023},
  publisher={Nature Publishing Group UK London}
}

@article{studier2022spectral,
  title={Spectral organ fingerprints for machine learning-based intraoperative tissue classification with hyperspectral imaging in a porcine model},
  author={Studier-Fischer, Alexander and Seidlitz, Silvia and Sellner, Jan and {\"O}zdemir, Berkin and Wiesenfarth, Manuel and Ayala, Leonardo and Odenthal, Jan and Kn{\"o}dler, Samuel and Kowalewski, Karl Friedrich and Haney, Caelan Max and others},
  journal={Scientific Reports},
  volume={12},
  number={1},
  pages={11028},
  year={2022},
  publisher={Nature Publishing Group UK London}
}

@article{seidlitz2022robust,
  title={Robust deep learning-based semantic organ segmentation in hyperspectral images},
  author={Seidlitz, Silvia and Sellner, Jan and Odenthal, Jan and {\"O}zdemir, Berkin and Studier-Fischer, Alexander and Kn{\"o}dler, Samuel and Ayala, Leonardo and Adler, Tim J and Kenngott, Hannes G and Tizabi, Minu and others},
  journal={Medical Image Analysis},
  volume={80},
  pages={102488},
  year={2022},
  publisher={Elsevier}
}

@article{garcia2023hyperspectral,
  title={Hyperspectral image segmentation: a preliminary study on the Oral and Dental Spectral Image Database (ODSI-DB)},
  author={Garcia Peraza Herrera, Luis C and Horgan, Conor and Ourselin, Sebastien and Ebner, Michael and Vercauteren, Tom},
  journal={Computer Methods in Biomechanics and Biomedical Engineering: Imaging \& Visualization},
  volume={11},
  number={4},
  pages={1290--1298},
  year={2023},
  publisher={Taylor \& Francis}
}

@article{hyttinen2020oral,
  title={Oral and dental spectral image database—odsi-db},
  author={Hyttinen, Joni and F{\"a}lt, Pauli and J{\"a}sberg, Heli and Kullaa, Arja and Hauta-Kasari, Markku},
  journal={Applied Sciences},
  volume={10},
  number={20},
  pages={7246},
  year={2020},
  publisher={MDPI}
}

@article{fabelo2019vivo,
  title={In-vivo hyperspectral human brain image database for brain cancer detection},
  author={Fabelo, Himar and Ortega, Samuel and Szolna, Adam and Bulters, Diederik and Pi{\~n}eiro, Juan F and Kabwama, Silvester and JO'Shanahan, Aruma and Bulstrode, Harry and Bisshopp, Sara and Kiran, B Ravi and others},
  journal={IEEe Access},
  volume={7},
  pages={39098--39116},
  year={2019},
  publisher={IEEE}
}

@article{martin2025slimbrain,
  title={SLIMBRAIN database: A multimodal image database of in vivo human brains for tumour detection},
  author={Mart{\'\i}n-P{\'e}rez, Alberto and Villa, Manuel and Rosa Olmeda, Gonzalo and Sancho, Jaime and Vazquez, Guillermo and Urbanos, Gemma and Martinez de Ternero, Alejandro and Chavarr{\'\i}as, Miguel and Jimenez-Roldan, Luis and Perez-Nu{\~n}ez, Angel and others},
  journal={Scientific Data},
  volume={12},
  number={1},
  pages={836},
  year={2025},
  publisher={Nature Publishing Group UK London}
}

@article{kopriva2025hyperspectral,
  title={A hyperspectral imaging dataset and Grassmann manifold method for intraoperative pixel-wise classification of metastatic colon cancer in the liver},
  author={Kopriva, Ivica and Sitnik, Dario and Dion-Bertrand, Laura-Isabelle and Peri{\v{s}}a, Marija Milkovi{\'c} and Pa{\v{c}}i{\'c}, Arijana and Had{\v{z}}ija, Mirko and Had{\v{z}}ija, Marijana Popovi{\'c}},
  journal={Computers in biology and medicine},
  volume={196},
  pages={110841},
  year={2025},
  publisher={Elsevier}
}

@article{aloupogianni2022hyperspectral,
  title={Hyperspectral imaging for tumor segmentation on pigmented skin lesions},
  author={Aloupogianni, Eleni and Ichimura, Takaya and Hamada, Mei and Ishikawa, Masahiro and Murakami, Takuo and Sasaki, Atsushi and Nakamura, Koichiro and Kobayashi, Naoki and Obi, Takashi},
  journal={Journal of Biomedical Optics},
  volume={27},
  number={10},
  pages={106007--106007},
  year={2022},
  publisher={Society of Photo-Optical Instrumentation Engineers}
}

@article{pathak2023spectral,
  title={Spectral similarity measures for in vivo human tissue discrimination based on hyperspectral imaging},
  author={Pathak, Priya and Chalopin, Claire and Zick, Laura and K{\"o}hler, Hannes and Pfahl, Annekatrin and Rayes, Nada and Gockel, Ines and Neumuth, Thomas and Melzer, Andreas and Jansen-Winkeln, Boris and others},
  journal={Diagnostics},
  volume={13},
  number={2},
  pages={195},
  year={2023},
  publisher={MDPI}
}

@article{puustinen2023hyperspectral,
  title={Hyperspectral placenta dataset: Hyperspectral image acquisition, annotations, and processing of biological tissues in microsurgical training},
  author={Puustinen, Sami and Hyttinen, Joni and Elomaa, Antti-Pekka and Vrz{\'a}kov{\'a}, Hana},
  journal={Data in Brief},
  volume={50},
  pages={109526},
  year={2023},
  publisher={Elsevier}
}

@article{giannoni2024transportable,
  title={Transportable hyperspectral imaging setup based on fast, high-density spectral scanning for in situ quantitative biochemical mapping of fresh tissue biopsies},
  author={Giannoni, Luca and Marradi, Marta and Scibilia, Kevin and Ezhov, Ivan and Bonaudo, Camilla and Artemiou, Angelos and Toaha, Anam and Lange, Fr{\'e}d{\'e}ric and Caredda, Charly and Montcel, Bruno and others},
  journal={Journal of Biomedical Optics},
  volume={29},
  number={9},
  pages={093508--093508},
  year={2024},
  publisher={Society of Photo-Optical Instrumentation Engineers}
}

@inproceedings{ronneberger2015u,
  title={U-net: Convolutional networks for biomedical image segmentation},
  author={Ronneberger, Olaf and Fischer, Philipp and Brox, Thomas},
  booktitle={International Conference on Medical image computing and computer-assisted intervention},
  pages={234--241},
  year={2015},
  organization={Springer}
}

@inproceedings{myronenko20183d,
  title={3D MRI brain tumor segmentation using autoencoder regularization},
  author={Myronenko, Andriy},
  booktitle={International MICCAI brainlesion workshop},
  pages={311--320},
  year={2018},
  organization={Springer}
}

@inproceedings{hatamizadeh2022unetr,
  title={Unetr: Transformers for 3d medical image segmentation},
  author={Hatamizadeh, Ali and Tang, Yucheng and Nath, Vishwesh and Yang, Dong and Myronenko, Andriy and Landman, Bennett and Roth, Holger R and Xu, Daguang},
  booktitle={2022 IEEE/CVF winter conference on applications of computer vision (WACV)},
  pages={1748--1758},
  year={2022},
  organization={Ieee}
}

@inproceedings{tang2022self,
  title={Self-supervised pre-training of swin transformers for 3d medical image analysis},
  author={Tang, Yucheng and Yang, Dong and Li, Wenqi and Roth, Holger R and Landman, Bennett and Xu, Daguang and Nath, Vishwesh and Hatamizadeh, Ali},
  booktitle={2022 IEEE/CVF Conference on Computer Vision and Pattern Recognition (CVPR)},
  pages={20698--20708},
  year={2022},
  organization={IEEE}
}

@inproceedings{cciccek20163d,
  title={3D U-Net: learning dense volumetric segmentation from sparse annotation},
  author={{\c{C}}i{\c{c}}ek, {\"O}zg{\"u}n and Abdulkadir, Ahmed and Lienkamp, Soeren S and Brox, Thomas and Ronneberger, Olaf},
  booktitle={International conference on medical image computing and computer-assisted intervention},
  pages={424--432},
  year={2016},
  organization={Springer}
}

@inproceedings{milletari2016v,
  title={V-net: Fully convolutional neural networks for volumetric medical image segmentation},
  author={Milletari, Fausto and Navab, Nassir and Ahmadi, Seyed-Ahmad},
  booktitle={2016 fourth international conference on 3D vision (3DV)},
  pages={565--571},
  year={2016},
  organization={Ieee}
}

@article{halicek2020tumor,
  title={Tumor detection of the thyroid and salivary glands using hyperspectral imaging and deep learning},
  author={Halicek, Martin and Dormer, James D and Little, James V and Chen, Amy Y and Fei, Baowei},
  journal={Biomedical Optics Express},
  volume={11},
  number={3},
  pages={1383--1400},
  year={2020},
  publisher={Optical Society of America}
}
\end{document}